Imitation versus Innovation: What children can do that large language and
language-and-vision models cannot (yet)?

Eunice Yiu[1], Eliza Kosoy[1], & Alison Gopnik[1]

[1] Department of Psychology, University of California - Berkeley, 2121 Berkeley Way, Berkeley,
CA, 94704, U.S.A.

Corresponding author: Eunice Yiu, University of California - Berkeley, 2121 Berkeley Way,
Berkeley, CA, 94704, U.S.A.

Email: ey242@berkeley.edu

## Abstract

Much discussion about large language models and language-and-vision models has focused on whether these models are intelligent agents. We present an alternative perspective. We argue that these artificial intelligence models are cultural technologies that enhance cultural transmission in the modern world, and are efficient imitation engines. We explore what AI models can tell us about imitation and innovation by evaluating their capacity to design new tools and discover novel causal structures, and contrast their responses with those of human children. Our work serves as a first step in determining which particular representations and competences, as well as which kinds of knowledge or skill, can be derived from particular learning techniques and data. Critically, our findings suggest that machines may need more than large scale language and images to achieve what a child can do.

**Introduction**

Recently, large language and language-and-vision models, such as OpenAI's GPT and DALL-E, have sparked much interest and discussion. These systems are trained on an unprecedentedly large amount of image and text data and are built with billions of parameters. The systems generate novel text or images in response to prompts. Typically, they are pretrained with a relatively simple objective such as predicting the next item in a string of text correctly. In more recent systems, they are also fine-tuned through reinforcement learning methods – humans judge the texts and images the systems generate, and so further shape what the systems produce.

A common way of thinking about these systems is to treat them as individual agents, and then debate how intelligent those agents are. The phrase "an AI" rather than "AI" or "AI system," implying individual agency, is frequently used. Some have claimed that these models can tackle complex commands, perform abstract reasoning, such as inferring theory of mind (e.g., Kosinski, 2023) and demonstrate creativity (e.g., Summers-Stay et al., 2023).

We argue that this framing is wrong. Instead, we argue that the best way to think of these systems is as powerful new cultural technologies, analogous to earlier technologies like writing, print, libraries, internet search and even language itself. Large language and vision models provide a new method for easy and effective access to the vast amount of text that others have written and images that others have shaped. These AI systems offer a new means for cultural production and evolution, allowing information to be passed efficiently from one group of people to another (Bolin, 2012; Boyd & Richerson, 1988; Henrich, 2018).

Furthermore, we argue that large language and vision models provide us with an opportunity to discover which representations and cognitive capacities, in general, human or artificial, can be acquired purely through cultural transmission itself and which require independent contact with the external world. One central question in cognitive science is how much the meaning of a word, an object or a concept can be learned from the distribution of disembodied and amodal symbols in language (Grand et al., 2022; Landauer & Dumais, 1997; Piantadosi, 2023), and how much meaning depends on grounded perceptual and motor interactions with the world (Barsalou, 2008; Gibson, 1979).

This contrast extends beyond perceptual and motor representations themselves. For example, the kinds of causal representations that are embodied in theories, either formal scientific theories or intuitive theories are the result of truth-seeking epistemic processes (e.g., Gopnik & Wellman, 2012; Harris et al., 2018). They are evaluated with respect to an external world and make predictions about and shape actions in that world. New evidence from that world can radically revise them. Representations like these are designed to solve "the inverse problem" – the problem of reconstructing the structure of a novel, changing, external world from the data that we receive from that world. Although such representations may be very abstract, as in scientific theories, they ultimately depend on perceptual and motor abilities – they depend on being able to perceive the world and act on it in new ways.

These truth-seeking epistemic processes contrast with the processes that allow faithful transmission of representations from one agent to another, regardless of the relation between those representations and the external world. Such transmission may be crucial for abilities like language learning and social coordination. There is considerable evidence that mechanisms for this kind of faithful transmission are in place early in development and play a particularly important role in human cognition and culture (Meltzoff & Moore, 1977). LLM's enable and facilitate this kind of transmission.

This contrast between transmission and truth is in turn closely related to the imitation/innovation contrast in discussions of cultural evolution (Tomasello et al., 1993; Legare and Nielsen, 2015, Boyd & Richerson, 1988; Henrich, 2018). Cultural evolution depends on the balance between these two different kinds of cognitive mechanisms. Imitation allows the transmission of knowledge or skill from one agent to another (Boyd et al., 2011; Henrich, 2016). Innovation produces novel knowledge or skill through contact with a changing world (Derex, 2022). Imitation means that each individual agent does not have to innovate – they can take advantage of the cognitive discoveries of others. But imitation by itself would be useless if some agents did not also have the capacity to innovate. It is the combination of the two that allows cultural and technological progress.

However, in any given case, it may be difficult to determine which kinds of cognitive mechanisms produced a particular kind of representation or behavior, knowledge or skill. For example, my answer to an exam question in school might simply reflect the fact that I have remembered what I was taught. Or it might indicate that I have knowledge that would allow me to make novel predictions about or perform novel actions on the external world.  Probing the responses of large models may give us a tool to help answer that question – at least, in principle. If large models can reproduce particular competencies, for example, producing grammatical text in response to a prompt, that suggests that those abilities can be developed through imitation – extracting existing knowledge encoded in the minds of others. If not, that suggests that these capacities may require innovation – extracting knowledge from the external world.

In this paper, we explore what state-of-the-art large language and language-and-vision models can contribute to our understanding of imitation and innovation. We contrast the performance of models trained on a large corpus of text data, or text and image data, with that of children.

## Large language and language-and-vision models as imitation engines.

Imitation refers to the behavior of copying or reproducing features or strategies underlying a model's behavior (Heyes, 2001; Tomasello, 1990). By observing and imitating others, individuals acquire the skills, knowledge, and conventions that are essential to effectively participate in their cultural groups, promoting cultural continuity over time. An assortment of technological innovations such as writing, print, internet search and we would argue, LLM's, have made this imitation much more effective over time.

Moreover, cultural technologies not only allow access to information, but they also codify, summarize, and organize that information in ways that enable and facilitate transmission. Language itself works by compressing information into a digital code. Writing and print similarly abstract and simplify from the richer information stream of spoken language, at the same time that they allow wider temporal and spatial access to that information. Print, in addition, allows many people to receive the same information at the same time, and this is, of course, highly amplified by the internet. At the same time, devices such as indexes, catalogs, and libraries, and more recently, Wiki's and algorithmic search engines, allow humans to quickly find relevant text and images and use those texts and images as a springboard to generate additional text and images.

Deep learning models trained on large datasets today excel at imitation in a way that far outstrips earlier technologies and so represent a new phase in the history of cultural technologies. Large language models such as Anthropic's Claude and OpenAI's GPT can use the statistical patterns in the text in their training sets to generate a variety of new text, from emails and essays to computer programs and songs. GPT-3 can imitate both natural human language patterns and particular styles of writing close to perfectly. It arguably does this better than many people (Zhang and Li, 2021). Strikingly and surprisingly, the syntactic structure of the language produced by these systems is accurate. There is some evidence that large language models can even grasp language in more abstract ways and imitate human figurative language understanding (e.g., Jeretic et al., 2020; Stowe et al., 2022). This suggests that finding patterns in large amounts of human text may be enough to pick up many features of language, independent of any knowledge about the external world. Similarly, large diffusion-based-text-to-image vision and language models such as DALLE-2 impressively create images that imitate a plethora of existing artistic styles and concepts (Fig. 1) from textual descriptions.

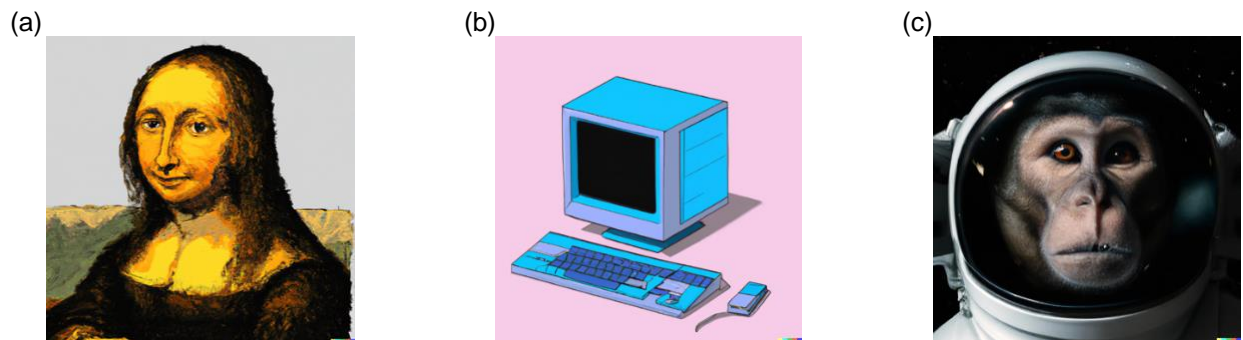

(a)  (b)  (c)

**Fig. 1.** Sample output images from DALLE-2 for the captions (a) "Mona Lisa painted by Vincent van Gogh," (b) "a computer from the 90s in the style of vaporwave" and (c) "high quality photo of a monkey astronaut."

In turn, this raises the possibility that human children learn features of language or images in a similar way. In particular, this discovery has interesting connections to the large body of empirical literature showing that infants are sensitive to the statistical structure of linguistic strings and visual images from a very young age (e.g., Kirkham, Slemmer & Johnson, 2002; Saffran, Aslin & Newport, 1996). The LLMs suggest that this may enable much more powerful kinds of learning than we might have thought, such as the ability to learn complex syntax.

On the other hand, although these systems, like children, are skilled imitators, their imitation may differ from that of children in important ways. There are debates in the developmental literature about how much childhood imitation simply reflects faithful cultural transmission (as in the phenomenon of over-imitation, where children reproduce unnecessary details of an action) and how much it is shaped by and in the service of broader truth-seeking processes such as understanding the goals and intentions of others. Children can meaningfully decompose observed visual and motor patterns in relation to the agent, target object, movement path and other salient features of events (Bekkering, Wohlschlager & Gattis, 2000; Gergely, Bekkering & Király, 2002). Moreover, children distinctively copy intentional actions (Meltzoff, 1995), discarding apparently failed attempts, mistakes, and causally inefficient actions (Schulz, Hooppell and Jenkins, 2008; Buchsbaum et al., 2011) when they seek to learn skills from observing other people (Over & Carpenter, 2013). While the imitative behavior of large language and vision models can be viewed as the abstract mapping of one pattern to another, human imitation appears to be mediated by goal representation and the understanding of causal structure from a young age. It would be interesting to see if large models also replicate these features of human imitation.

**Can large language and language-and-vision models innovate?**

**I.     Can LLM's design new tools?**

The most ancient representative of the human genus is called *Homo habilis* ("handy man") due to their ability to design and use novel stone tools. Tool use is one of the best examples of the advantages of cultural transmission, and of the balance between imitation and innovation. Imitation allows a novice to observe a model and reproduce their actions to bring about a particular outcome, even without really understanding the causal properties of the tool. Techniques like "behavior cloning" in AI and robotics use a similar approach.

Again however, the ability to imitate existing tools depends on the parallel ability to design new tools. Tool innovation is an indispensable part of human lives, and it has also been observed in a variety of nonhuman animals such as crows (Von Bayern et al., 2009) and chimpanzees (Whiten, Horner & de Waal, 2005). Tool innovation has often been taken to be a distinctive mark of intelligence in biological systems (Emery & Clayton, 2004; Reader & Laland, 2002).

Tool use can then be an interesting point of comparison between large models and children. We might predict that these models will generate text and images that capture familiar tool uses – for example, predicting appropriately that a hammer should be used to bang in a nail. However, these systems might have more difficulty producing the right responses for tool innovation, which depends on discovering and using new causal properties and affordances. We might, however, also wonder whether young children can themselves perform this kind of innovation, or whether it depends on explicit instruction and experience. In an ongoing study of tool innovation (Yiu & Gopnik, *in press*), we have investigated whether human children and adults can insightfully use familiar objects in new ways to accomplish particular outcomes, and compared the results to the output of large deep learning models such as GPT-3 and DALLE-2.

Tool innovation can involve designing new tools, but it can also refer to using old tools in new ways to solve novel problems (Rawlings and Legare, 2021). Our experiment examines the latter type of tool innovation. Physically building a new tool from scratch and then executing a series of actions that lead to a desired goal is a difficult task for young children (Beck et al., 2011). So instead, we study the ability to recognize new functions in everyday objects and to select appropriate object substitutes in the absence of typical tools to solve various physical tasks.

Our study has two components: an "imitation" component and an "innovation" component. In the innovation part of the study, we present a series of problems in which a goal has to be executed in the absence of the typical tool (e.g., holding a hot cup of coffee in the absence of a mug sleeve). We then provide alternative objects for participants to select: (1) an object that is more superficially similar to the typical tool, that is the mug sleeve, and is associated with it but is not functionally relevant to the context (e.g., a coffee lid), (2) an object that is superficially dissimilar but has the same affordances and causal properties as the typical tool (e.g., a plant pot), and (3) a totally irrelevant object (e.g., a succulent). In the imitation part of the study, we present the same sets of objects but ask participants to select which of the object options would "go best" with the typical tool (e.g., a mug sleeve and a coffee lid should go together).

So far, we have found that both children aged three to seven years old ($n = 42$, $M_{age} = 5.71$ years, $SD = 1.24$) and adults ($n = 30$, $M_{age} = 27.80$ years, $SD = 5.54$) can recognize common superficial relationships between objects when they are asked which objects should go together ($M_{children} = 88.4\%$, $SE_{children} = 2.82\%$; $M_{adults} = 84.9\%$, $SE_{adults} = 3.07\%$). But they can also discover new functions in everyday objects to solve novel physical problems and so select the superficially unrelated but functionally relevant object. ($M_{children} = 85.2\%$, $SE_{children} = 3.17\%$; $M_{adults} = 95.7\%$, $SE_{adults} = 1.04\%$).

Using exactly the same questions that we used to test our human participants, we queried OpenAI's gpt-3.5-turbo and text-davinci-003 models, Anthropic's Claude, and Google's FLAN-T5 (XXL) and BigScience's Bloomz. As we predicted we found that these large language models are almost as capable of identifying superficial commonalities between objects as humans are. They are sensitive to the associations between the objects, and they excel at our imitation tasks – they respond that the coffee lid goes with the hot coffee. However, they are much less capable than both adults and children when they are asked to select a novel functional tool to solve a problem – they again choose the coffee lid rather than the plant pot to hold the hot coffee. This suggests that simply learning from large amounts of existing language may not be sufficient to achieve grounded tool innovation. Discovering novel functions in everyday tools is not about finding the statistically nearest neighbor from lexical co-occurrence patterns. Rather, it is about appreciating the more abstract functional analogies and causal relationships between objects that do not necessarily belong to the same category or are associated in text. Compared to humans, large language models are not as successful at this type of innovation task. On the other hand, they excel at generating responses that simply demand some abstraction from existing knowledge.

Perhaps this reflects the fact that these tasks involve visual and spatial information rather than just text. To explore this possibility, we tested whether large vision and language models such as DALLE-2 can depict how common physical problems like those in our experiment can be solved. We looked both at how these systems represent the use of typical tools versus the innovative tools that our human participants selected. DALLE-2 was reasonably good at producing images that capture how typical tools are commonly used (Fig. 2(a), (c), (e)). However, when a more unconventional tool was mentioned in the caption, the system could not appropriately piece together how these objects can be applied to achieve the described goals (Fig. 2(b), (d), (f)).

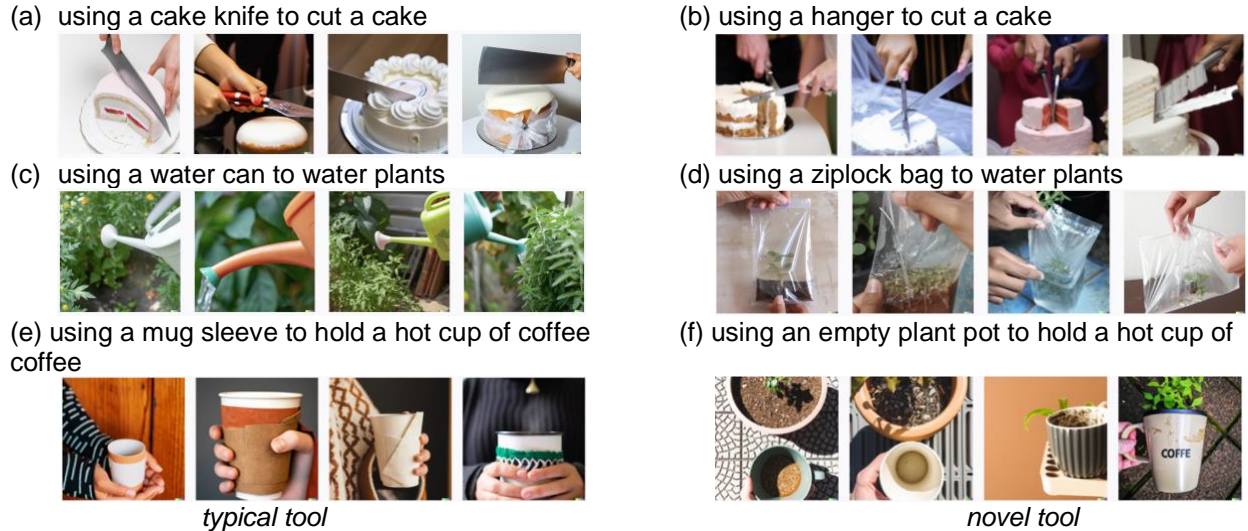

(a) using a cake knife to cut a cake

(b) using a hanger to cut a cake

(c) using a water can to water plants

(d) using a ziplock bag to water plants

(e) using a mug sleeve to hold a hot cup of coffee

(f) using an empty plant pot to hold a hot cup of coffee

*typical tool*                               *novel tool*

**Fig. 2.** Examples of outputs by DALLE-2 for solving physical problems using typical tools and novel tools.

All in all, creativity and innovation refer to the production of ideas and products that are both novel and useful given the constraints of a certain context (Barron, 1955; Runco & Jaeger, 2012). This does not seem to be consistently achievable in current large language models yet. This reflects the lack of relation between the model representations and the external world.

**II.    Can LLM's discover novel causal relationships and use them to design interventions?**

Designing novel tools depends on being able to infer a novel causal relationship, such as the insulating relationship between the plant pot and the coffee cup. A substantial literature shows that even very young children excel at discovering such relationships. Information about causal structure can be conveyed through imitation and cultural transmission. In fact, from a very young age, even infants will reproduce an action they have observed in order to bring about an effect (Waismeyer, Meltzoff & Gopnik, 2015). However, very young children can also infer novel causal structure by observing complex statistical relations among events, and ,most significantly, by acting on the world themselves to bring about effects like a scientist performing experiments (Cook, Goodman & Schulz, 2011; Gopnik et al., 2004, 2017; Gopnik &

Tenenbaum, 2007; Schulz, Bonawitz & Griffiths, 2007). Causal discovery is a particularly good example of a cognitive process that is directed at solving an inverse problem and discovering new truths through perception and action.

In another line of research (Kosoy et al., 2022, in prep) we have explored whether LLM's and other AI models can discover and use novel causal structure. In these studies we use a virtual "blicket detector" – a machine that lights up and plays music when you put some objects on it but not others. The blicket detector can work on different abstract principals or "over-hypotheses", individual blocks may make it go or you may need a combination of blocks to do so. An over-hypothesis refers to an abstract principal that reduces a hypothesis space at a less abstract level (Kemp, Performs & Tenenbaum, 2007), and a causal over-hypothesis refers to transferable abstract hypotheses about sets of causal relationships (Kosoy et al., 2022). If you know that it takes two blocks to make the machine go, you will generate different specific hypotheses about which blocks are blickets.

The blicket detector tasks intentionally involve a new artifact, described with new words, so that the participants cannot easily use past culturally transmitted information, such as the fact that flicking a light switch makes a bulb go on. In these experiments, we simply ask children to figure out how the machines work and allow them to freely explore and act to solve the task and determine which blocks are blickets. Even four-year-old children spontaneously acted on the systems and discovered their structure – they figured out which ones were blickets and used them to make the machine go.

We then gave a variety of LLM's, including OpenAI's GPT, Google's PaLM and most recently LaMDA the same data that the children produced, described in language (e.g., "I put the blue one and the red one on the machine and the machine lit up" and prompted the systems to answer questions about the causal structure of the machine (e.g., Is the red one a blicket?). Machine learning systems struggled to model, understand and extract causal over-hypotheses from their data. Young children learned novel causal over-hypotheses from only a handful of observations, including the outcome of their own experimental interventions, and applied the learned structure to novel situations. In contrast, large language models and vision-and-language models, as well as both deep reinforcement learning algorithms and behavior cloning struggled to figure out the relevant causal structures. This is consistent with other recent studies: LLM's produce the correct text in cases like causal vignettes, where the patterns are available in the training data, but often fail when they are asked to make inferences that involve novel events or relations (e.g., Binz and Schulz, 2022; Ullman, 2023), even when these involve superficially slight changes to the training data.

**Challenges of Studying Large Language and Language-and-Vision Models: The Questions Left Unanswered**

It is difficult to escape the language of individual agency, for example, to ask whether an LLM can or cannot innovate or solve a causal problem, or even can or cannot be sentient or intelligent. A great deal of the discussion about AI has this character. But we emphasize again that the point of this work is neither to decide whether or not LLM's are intelligent agents, nor to present some crucial comparative gotcha test that would determine the answer to such questions. Instead, the research projects we have briefly described here are a first step in determining which representations and competences, as well as which kinds of knowledge or skill, can be derived from which learning techniques and data. Which kinds of knowledge can be extracted from large bodies of text and images, and which depend on actively seeking the truth about an external world?

We think that there is a great deal of scope for research that uses developmental psychology techniques to investigate AI systems and vice-versa. Developmentalists have long realized that superficially similar behaviors can have very different psychological origins and can be the result of very different learning techniques and data. As a result, we have put considerable methodological energy into trying to solve this problem. A particular conversation with a child, however compelling, is just the start of a proper research program including novel, carefully controlled tests like our tests of tool innovation and causal inference. The conversation may reflect knowledge that has come through imitation, statistical pattern recognition, reinforcement from adults or conceptual understanding and the job of the developmental psychologist is to distinguish these possibilities. This should also be true of our assessments of AI systems.

At the same time AI systems have their own properties that need to be considered when we compare their output to that of humans. These can sometimes be problematic, for example, once a particular cognitive test is explicitly described in internet text it then becomes part of the system's training sample – we found that in some cases the systems referred to our earlier published blicket detector papers as the source for their answers! The more recent versions of GPT, GPT-4 and GPT-3.5, have also been fine-tuned through reinforcement learning from human feedback. This also raises problems, human reinforcement may be opaque and variable, and may simply edit out the most obvious mistakes and errors. But on the other hand, these systems also have the advantage that we know more about their data and learning techniques than we do about those of human children. For example, we know that the data for GPT systems is internet text and that the training function involves predicting new text from earlier text. We know that large language models and language-and-vision models are built on deep neural networks and trained on immense amounts of unlabeled text or text-image pairings.

These kinds of techniques may indeed contribute to some kinds of human learning as well. Children do learn through cultural transmission and statistical generalizations from data. But human children also learn in very different ways. Though we don't know the details of the child's learning algorithms or data, we do know that, unlike large language and language-and-vision models, they are curious, active, self-supervised and intrinsically motivated. They are capable of extracting novel and abstract structures from the environment beyond statistical patterns, spontaneously making over-hypotheses and generalizations, and applying these insights to new situations. A child's learning is also embodied in the physical world; changes in the brain, body, environment, and experiences promote behavioral flexibility and multimodal exploration, which leads to cascades of development in various psychological domains (Adolph and Hoch, 2019; Iverson, 2021).

Since performance in large deep learning models has been steadily improving with increasing model size on various tasks, some have advocated that simply scaling up language models could allow task-agnostic, few-shot performance (e.g., Brown et al., 2020). But a child does not interact with the world better by increasing their brain capacity. Is building the tallest tower the ultimate way to reach the moon? Putting scale aside, what are the mechanisms that allow us humans to be effective and creative learners? What in a child's "training data" and learning capacities is critically effective and different from that of LLM's? Can we design new AI systems that use active, self-motivated, exploration of the real external world as children do? And what we might expect the capacities of such systems to be? Comparing these systems in a detailed and rigorous way can provide important new insights about both natural and artificial intelligence.

**Conclusion**

Large language models such as ChatGPT are valuable cultural technologies. They can imitate millions of human writers, summarizing long texts, translating between languages, answering questions and coding programs. Imitative learning is critical for promoting and preserving knowledge, artifacts, and practices faithfully within social groups. Moreover, changes in cultural technologies can have transformative effects on human societies and cultures – for good or ill. There is a good argument that the initial development of printing technology contributed to the Protestant reformation. Later improvements in printing technology in the 18$^{th}$ century were responsible for both the best parts of the American Revolution and the worst parts of the French one (Darnton, 1982). Large language and language-and-vision models may well have equally transformative effects in the 21$^{st}$ century.

However, cultural evolution depends on a fine balance between imitation and innovation. There would be no progress without innovation, the ability to expand, create, change, abandon, evaluate and improve on existing knowledge and skills. Whether this means recasting existing knowledge in new ways or creating something entirely original, innovation challenges the status quo and questions the conventional wisdom that is the training corpus for artificially intelligent systems. Large language models can help us acquire information that is already known more efficiently, even though they are not innovators themselves. Moreover, accessing existing knowledge much more effectively can stimulate more innovation among humans and perhaps the development of more advanced AI. But ultimately, machines may need more than large scale language and images to match the achievements of every human child.